\documentclass[letterpaper]{article} 
\usepackage{aaai2026}  
\usepackage{times}  
\usepackage{helvet}  
\usepackage{courier}  
\usepackage[hyphens]{url}  
\usepackage{graphicx} 
\usepackage{natbib}  
\usepackage{caption} 
\usepackage{algorithm}

\usepackage{amsmath}
\usepackage{amssymb}
\usepackage{dsfont}
\usepackage{amsthm}

\usepackage{mathrsfs}
\usepackage{subcaption}
\usepackage{enumitem}
\usepackage{xcolor}
\usepackage{booktabs} 
\usepackage{multirow}

\usepackage{algpseudocode}

\newtheorem{prop}{PROPOSITION}

\usepackage{newfloat}
\usepackage{listings}
\DeclareCaptionStyle{ruled}{labelfont=normalfont,labelsep=colon,strut=off} 
\floatstyle{ruled}
\newfloat{listing}{tb}{lst}{}
\floatname{listing}{Listing}
\title{$\mathcal{IDK}$-$\mathcal{S}$: Incremental Distributional Kernel for Streaming Anomaly Detection}

\author {
    Yang Xu\textsuperscript{\rm 1,2}, 
    Yixiao Ma\textsuperscript{\rm 1,2}, 
    Kaifeng Zhang\textsuperscript{\rm 1,2}, 
    Zuliang Yang\textsuperscript{\rm 1,2}, 
    Kai Ming Ting\textsuperscript{\rm 1,2}\thanks{Corresponding author.}
}

\affiliations {
    \textsuperscript{\rm 1} State Key Laboratory for Novel Software Technology, Nanjing University, Nanjing, China\\
    \textsuperscript{\rm 2} School of Artificial Intelligence, Nanjing University, Nanjing, China\\
    \{xuyang, mayx, zhangkf2022, yangzl\}@lamda.nju.edu.cn, tingkm@nju.edu.cn
}

\begin{document}

\maketitle


\begin{abstract}

Anomaly detection on data streams presents significant challenges, requiring methods to maintain high detection accuracy among evolving distributions while ensuring real-time efficiency. Here we introduce $\mathcal{IDK}$-$\mathcal{S}$, a novel $\mathbf{I}$ncremental $\mathbf{D}$istributional $\mathbf{K}$ernel for $\mathbf{S}$treaming anomaly detection that effectively addresses these challenges by creating a new dynamic representation in the kernel mean embedding framework. The superiority of $\mathcal{IDK}$-$\mathcal{S}$ is attributed to two key innovations. First, it inherits the strengths of the Isolation Distributional Kernel, an offline detector that has demonstrated significant performance advantages over foundational methods like Isolation Forest and Local Outlier Factor due to the use of a data-dependent kernel. Second, it adopts a lightweight incremental update mechanism that significantly reduces computational overhead compared to the naive baseline strategy of performing a full model retraining. This is achieved without compromising detection accuracy, a claim supported by its statistical equivalence to the full retrained model. Our extensive experiments on thirteen benchmarks demonstrate that $\mathcal{IDK}$-$\mathcal{S}$ achieves superior detection accuracy while operating substantially faster, in many cases by an order of magnitude, than existing state-of-the-art methods.

\end{abstract}


\section{Introduction}

Anomaly detection, also known as outlier detection, is an essential task in various domains, identifying unusual patterns that do not conform to expected behavior~\cite{chandola2009anomaly}. This technique is widely applied in many areas, such as financial fraud detection~\cite{hilal2022financial}, network security~\cite{ahmed2016survey}, and industrial fault monitoring~\cite{stojanovic2016big}, where anomalous behaviors can signify important and often costly events. 

With the exponential growth of data generation, many modern applications in anomaly detection require processing data not as static datasets, but as continuous data streams. This is often referred to as online anomaly detection which imposes strict constraints: methods must process each data instance upon arrival with limited memory and must be highly adaptable to evolving data distributions, a phenomenon known as concept drift~\cite{cao2025revisiting}. In such dynamic environments, offline anomaly detection solutions are rendered ineffective, as their static models quickly become obsolete and fail to represent the current data normality. Although a straightforward approach is to continuously retrain the model as new data emerge, this strategy incurs substantial computational overhead and latency, making it impractical for high-throughput, real-time applications.

The prevailing strategy for developing streaming anomaly detectors is to adapt an established offline method, modifying its model to handle data incrementally and thus avoid the repeated high cost of reconstructing a full model~\cite{lu2023review}. Existing state-of-the-art methods are built upon a few foundational offline detectors, including Isolation Forest (iForest)~\cite{liu2008isolation}, Local Outlier Factor (LOF)~\cite{breunig2000lof}, and k-Nearest Neighbors (kNN) detector~\cite{ramaswamy2000efficient}. Recent prominent examples of this trend include online Isolation Forest (oIFOR)~\cite{leveni2024online} and streaming Isolation Forest (SiForest)~\cite{Liu2025Streaming}, both of which extend iForest to the streaming context. For example, oIFOR implements an efficient incremental update by modeling the data distribution with an ensemble of dynamic, multi-resolution histograms, which adapt their structure to learn and forget instances from a sliding buffer. 
This adaptation-based design is popular as it leverages the well-understood properties of established offline methods. However, an often overlooked consequence of this design is that the effectiveness of such a streaming algorithm depends on the capabilities of its base detector. This means that not only the strengths of its base detector are inherited, but so are its weaknesses. 
For example, streaming methods built upon iForest are likely to inherit its weaknesses with varied densities and local anomalies~\cite{aryal2014improving,bandaragoda2014efficient}, while LOF-based methods are typically sensitive to its $k$ parameter setting in different data sizes~\cite{ting2021ikde}. These inherent weaknesses, inherited from their offline counterparts, create a performance ceiling that current streaming detectors struggle to break.

\begin{figure}[t]
    \centering
    \begin{subfigure}[b]{1\columnwidth}
        \includegraphics[width=\linewidth]{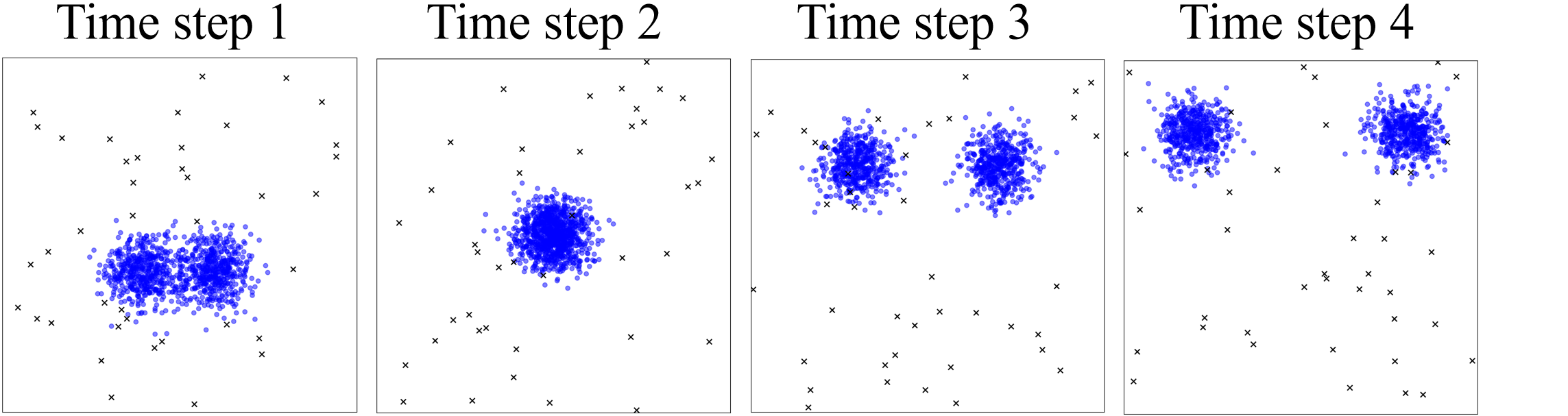}
        \caption{Data stream over four time steps.}
        \label{fig:s1}
    \end{subfigure}
    \hfill 
    \begin{subfigure}[b]{1\columnwidth}
        \includegraphics[width=\linewidth]{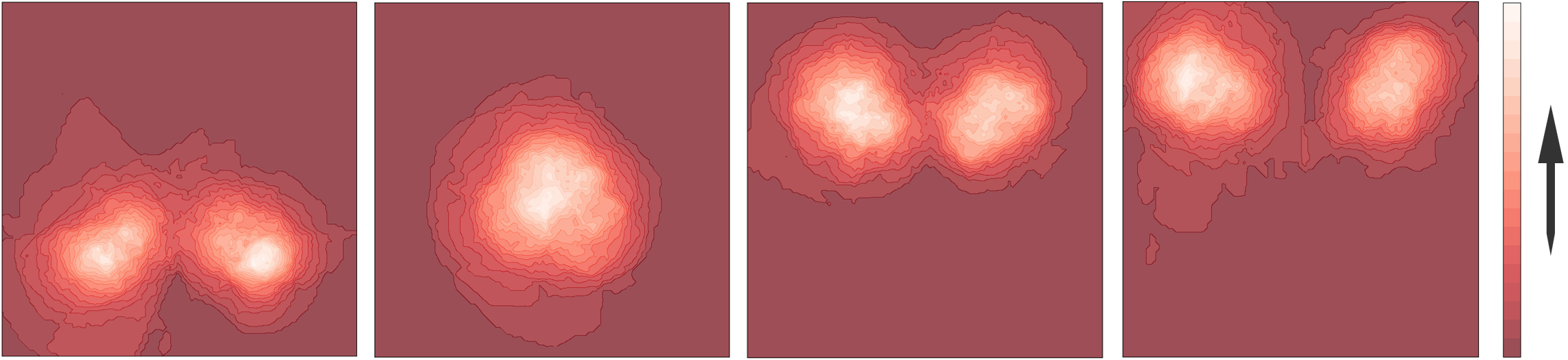}
        \caption{The distribution of normal scores over four time steps.}
        \label{fig:s2}
    \end{subfigure}
    \caption{An illustration of $\mathcal{IDK}$-$\mathcal{S}$'s adaptivity to concept drift. (a) A data stream where the normal distribution (blue clusters) shifts over four time steps. (b) The corresponding normal score distribution estimated by $\mathcal{IDK}$-$\mathcal{S}$. The heatmap (brighter areas indicate higher normal scores, i.e., higher normality) dynamically follows the evolving normality, demonstrating the model's ability to maintain accurate detection in a non-stationary online environment.\\
    }
    \label{fig:f1}
\end{figure}

Isolation Distributional Kernel (IDK)~\cite{ting2021isolation} is an anomaly detector based on kernel mean embedding~\cite{muandet2017kernel}. IDK maps a given static dataset to a single point in a Hilbert space, which represents  the data distribution of the dataset. To  score any test instance, its similarity w.r.t. the data distribution of the entire dataset is computed as a dot product in Hilbert space. The instances having low similarity are identified as anomalies. The superiority of IDK over other detectors derives from its use of a data-dependent kernel~\cite{ting2021isolation}. Specifically, the data-dependent property of the kernel denotes that any two instances, with equal inter-instance distance, are more similar when they are located in a sparse data region than in a dense one. This characteristic is crucial for successful anomaly detection in datasets having varied densities and local anomalies, a known challenge for many traditional detectors~\cite{boukerche2020outlier}. Furthermore, its use of an exact, finite-dimensional feature map ensures highly efficient, linear-time computation. Empirical studies have shown that IDK exhibits significantly superior performance compared to traditional offline detectors such as iForest, LOF, and kNN-based methods~\cite{ting2021ikde}. In a recent survey, the retraining-based IDK model is found to have superior performance to state-of-the-art methods for streaming anomaly detection~\cite{cao2025revisiting}. However, this retraining-based model requires a high time cost, making it impractical for potentially infinite data streams.

The demonstrated superiority of the IDK as an offline detector motivates our work to adapt this model for streaming anomaly detection. Here we propose $\mathcal{IDK}$-$\mathcal{S}$, a novel $\mathbf{I}$ncremental $\mathbf{D}$istributional $\mathbf{K}$ernel for $\mathbf{S}$treaming anomaly detection. $\mathcal{IDK}$-$\mathcal{S}$ uses a simple yet effective incremental update mechanism that adapts incrementally to newly arriving data. 
The retraining-based IDK ~\cite{cao2025revisiting} must rebuild the entire set of data-dependent partitions in IDK that creates the data-dependent property. 
Instead, our method updates the data-dependent partitions by 
continuously replacing the obsolete partitions with new ones generated from new arriving data, ensuring that the model remains aligned with the latest data distribution. Figure~\ref{fig:f1} illustrates the adaptive capability of our method with a toy streaming data (the \texttt{TwoCluster} dataset used in the experiments section). Normal instances are depicted as dense blue clusters of instances, while anomalies are depicted as sparsely distributed instances. As the distribution of normal instances changes over time (i.e., concept drift), our proposed method effectively updates its estimated scores for the evolving data stream, as shown on the heatmaps. 

In short, we highlight a critical yet often overlooked fact: \textit{the performance ceiling of  each existing streaming detector is fundamentally constrained by its offline counterpart.} To break through this ceiling, a more powerful base detector is essential. Motivated by this, our work makes the following contributions:
\begin{itemize}
    \item Proposing $\mathcal{IDK}$-$\mathcal{S}$ that adapts the superior IDK to evolving data streams via a new incremental partition-update mechanism; thus addressing the performance limitations of methods based on weaker offline detectors.
    \item Demonstrating that $\mathcal{IDK}$-$\mathcal{S}$ reduces the theoretical time complexity from $\mathcal{O}(\omega\psi t)$ of the retraining-based IDK model to $\mathcal{O}(l\psi t)$, where $\omega$ and $l$ are the data window size and the update step size respectively, and $l \ll \omega$.
    \item Showing the superiority of $\mathcal{IDK}$-$\mathcal{S}$ for both high detection accuracy and efficiency compared to state-of-the-art methods through experiments on thirteen benchmarks, including stationary and non-stationary distributions.
\end{itemize}

\section{Preliminaries}

\subsection{Problem Formulation}

We consider the streaming anomaly detection task in a potentially infinite multivariate data stream. A data stream $\mathcal{S}$ can be defined as a sequence of instances $x_T\in \mathbb{R}^d$:
\begin{equation}
    \mathcal{S} = \{x_1,x_2,\dots,x_T,\dots\},
\end{equation}
where $T>1$ and $x_T$ is an instance generated from an underlying data distribution $\mathcal{P}_T$ at time instant $T$, i.e., $x_T\sim \mathcal{P}_T$. In a streaming context with concept drift, this distribution can evolve over time, i.e., $\mathcal{P}_T$ may differ from $\mathcal{P}_{T+1}$.
In line with the existing online learning setting, we assume that the stream is processed in sliding windows. At any time step $i$, we have access to a finite data window $\mathbf{X}_i\subset\mathcal{S}$ of size $\omega$, which contains the most recent instances from the stream. An online detector must score each instance in $\mathbf{X}_i$, and its model is updated with $\mathbf{X}_i$.

Let $\mathcal{P}_N$ denote the distribution of normal instances and $\mathcal{P}_A$ denote the distribution of anomalies. 
Given the data stream $\mathcal{S}$ and any instance $x\in \mathcal{S}$, the goal is to learn a model $\mathcal{F}_i(x)$ to assign an anomaly score (or normal score) to  $x$, which identifies whether $x \sim \mathcal{P}_N$ or $x \sim \mathcal{P}_A$.
Following established studies on anomaly detection~\cite{liu2008isolation,leveni2024online}, we assume that anomalies are characterized by two properties: ``few" and ``different". For any instance $x\in \mathcal{S}$,
\begin{enumerate}[label=(\roman*)]
    \item The ``few" property implies that the probability that $x$ is generated by the anomaly distribution $\mathcal{P}_A$ is much lower than that  by the normal distribution $\mathcal{P}_N$, i.e., for $x \in \mathcal{S}$, 
    \begin{equation}
        P(x \sim \mathcal{P}_A) \ll P(x \sim \mathcal{P}_N).
    \end{equation}
    \item The ``different" property suggests that an anomalous instance $x_a$  will exhibit low similarity to the population of normal instances. More formally, its expected similarity to the normal data distribution $\mathcal{P}_N$ is significantly lower than that of a normal instance $x_n$, i.e., 
    \begin{equation}
        E_{y \sim \mathcal{P}_N}[\kappa(x_a, y)] \ll E_{y \sim \mathcal{P}_N}[\kappa(x_n, y)],
    \end{equation}
    where $\kappa$ is a similarity function.
\end{enumerate}
 
The challenge in streaming anomaly detection is to accurately and efficiently identify anomalies from unknown and evolving distributions under memory constraint, favoring incremental model updates over computationally expensive retraining.

\subsection{Background: Isolation Distributional Kernel}

IDK~\cite{ting2021ikde,ting2021isolation} is a powerful measure to quantify the similarity between two distributions, based on a specific data-dependent point kernel called Isolation Kernel (IK)~\cite{ting2018isolation}. It has been shown to improve the performance of various downstream tasks, such as density-based clustering~\cite{qin2019nearest}, t-SNE visualization~\cite{zhu2021improving}, and high-dimensional nearest neighbor search~\cite{ting2024possible,xu2025voronoi}, by mapping original data to a finite-dimensional feature map.

\begin{figure}[t]
    \centering
    \includegraphics[width=0.98\columnwidth]{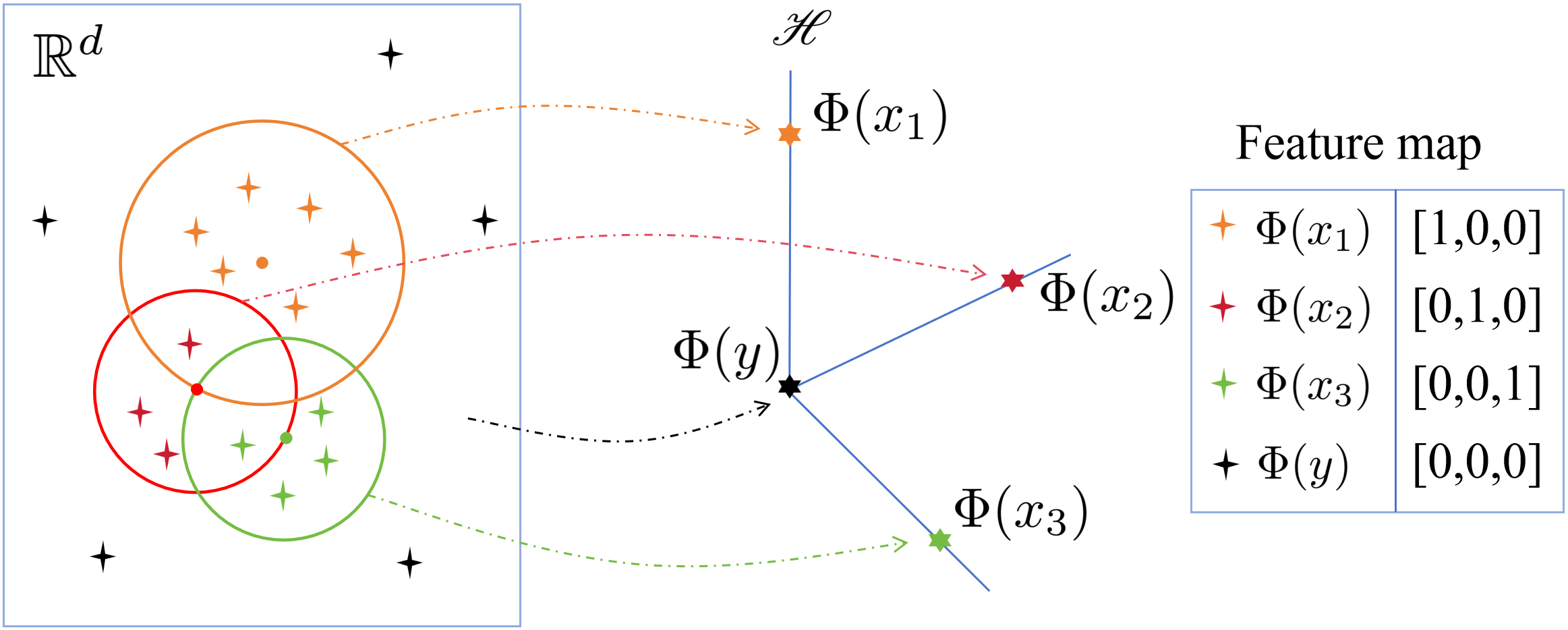} 
    \caption{An illustration of the feature map $\Phi$ of IK in a Hilbert space $\mathscr{H}$ with one partitioning ($t=1$) of three hyperspheres ($\psi=3$), each centred at a blue dot that is randomly selected from the given dataset $D$. When a point $x\in D$ falls into an overlapping region, it is regarded to be in the hypersphere whose centre is closer to $x$.}
    \label{fig:ik_map}
\end{figure}

IK defines similarity based on random data-driven partitions, which can be implemented with hyperspheres. 
This mechanism is illustrated in Figure~\ref{fig:ik_map}. It creates a random partitioning of the data space by first selecting a small subset of $\psi$ points from a static dataset $D$. Then, for each selected point, a hypersphere is centered on it with a radius determined by its distance to its nearest neighbor within that subset.
These $\psi$ hyperspheres, along with the remaining unbounded space, form a complete partition of $\mathbb{R}^d$.
IK uses an ensemble approach where this partitioning process is repeated $t$ times. This procedure yields its exact, finite-dimensional feature map, $\Phi(x) \in \{0,1\}^{\psi \times t}$, where each value indicates if $x$ falls into a specific hypersphere.

Then, based on the principle of KME~\cite{muandet2017kernel}, IDK allows the similarity between two distributions, $\mathcal{P}_X$ and $\mathcal{P}_Y$, to be computed efficiently as the inner product of their mean kernel vectors (i.e., feature mean maps). The IDK is thus defined as:
\begin{equation}
    \mathcal{K}_{\text{IDK}}(\mathcal{P}_X, \mathcal{P}_Y | D) = \frac{1}{t} \left\langle \widehat{\Phi}(\mathcal{P}_X), \widehat{\Phi}(\mathcal{P}_Y) \right\rangle,
\end{equation}
where the kernel mean map $\widehat{\Phi}(\mathcal{P}_X) = \frac{1}{|X|} \sum_{x \in X} \Phi(x)$ is the centroid of the feature vectors of all points sampled from the distribution $\mathcal{P}_X$.

In offline anomaly detection, IDK computes a single mean feature vector, $\widehat{\Phi}(\mathcal{P}_D)$, which represents the centroid of distribution of the static dataset $D$. The normality of any given instance $x$ is then measured by its similarity to the centroid of the overall data distribution. A high degree of similarity, computed via the inner product $\text{score}(x) = \frac{1}{t} \langle \Phi(x), \widehat{\Phi}(\mathcal{P}_D) \rangle$, results in a high normal score. Conversely, a low score indicates a poor fit with the overall data distribution, thus identifying a likely anomaly. The static nature of this design, however, presents a core challenge for dynamic data streams, which we address next.

\section{Methodology}
\label{sec:method}

\subsection{Insight: Incremental Distributional Kernel}
\label{sec:inc_kernel}

\begin{figure*}[t]
    \centering
    \begin{subfigure}[b]{1\columnwidth}
        \includegraphics[width=0.96\linewidth]{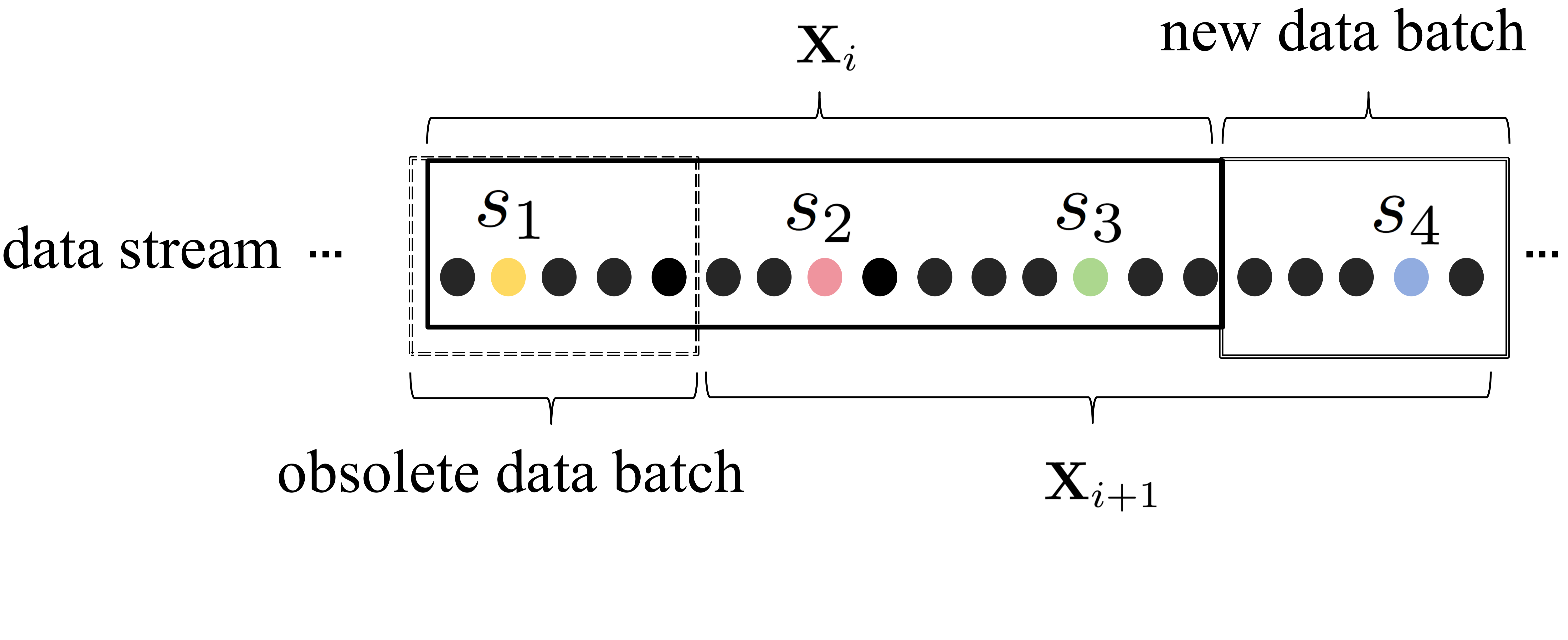}
        \caption{New data batch arrives.}
        \label{fig:window_slide}
    \end{subfigure}
    \hfill 
    \begin{subfigure}[b]{1\columnwidth}
        \includegraphics[width=0.96\linewidth]{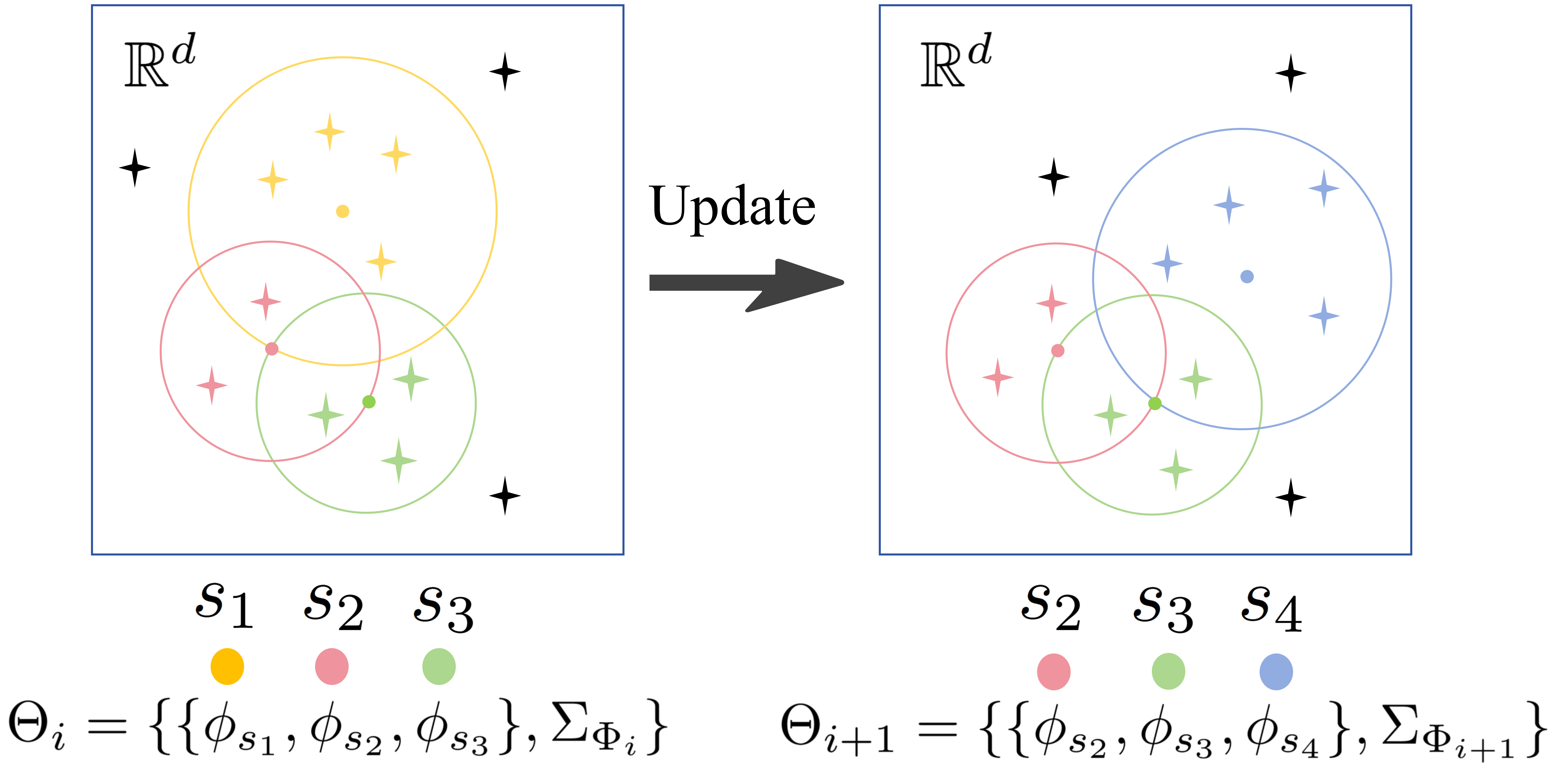}
        \caption{$\mathcal{IDK}$-$\mathcal{S}$ model updates.}
        \label{fig:model_update}
    \end{subfigure}
    \caption{An illustration of the incremental update mechanism of $\mathcal{IDK}$-$\mathcal{S}$. As the data slides from $\mathbf{X}_i$ to $\mathbf{X}_{i+1}$, the model updates by discarding the partition tied to an obsolete sample ($s_1$, yellow) and generating a new partition from an incoming sample ($s_4$, blue). This process selectively modifies the set of hypersphere partitions, updating the model from $\Theta_i$ to $\Theta_{i+1}$.}
    \label{fig:idk_s}
\end{figure*}

The standard IDK is an offline method designed for static datasets. Its core data-dependent partitions are generated once from a fixed reference dataset and remain unchanged thereafter. When applied directly to a data stream where the underlying distribution evolves (i.e., concept drift), a model built on historical data quickly becomes obsolete. Consequently, the naive strategy for the standard IDK is to periodically discard the entire model and retrain it from scratch on a new window of recent data~\cite{cao2025revisiting}. Although straightforward, this retraining-based model is computationally expensive and impractical for real-time applications.

Our proposed incremental strategy is motivated by a key insight into IDK's structure: \textit{the model can be treated as an ensemble of weak detectors, where each hypersphere partition acts as one such detector, which contributes one value to the entire feature map.} A value of $1$ in this feature map signifies that  an instance is considered ``locally normal" by that specific partition, while a $0$ means that it falls outside. The final anomaly score is not determined by any single detector, but by the aggregation of the entire feature map via KME. 
The power of this perspective is that since each weak detector is represented by a hypersphere, generated from a sampled point, the overall model can be updated incrementally by only replacing the detectors associated with obsolete sampled points. Specifically, as the data window slides, we identify the small subset of weak detectors whose generating samples came from the obsolete data batch. We then replace obsolete detectors with new ones generated from the new data batch. As the number of replaced detectors is small, this update strategy is far more efficient than retraining the entire IDK model.
We term this incremental update mechanism the Incremental Distributional Kernel, which forms the core of our $\mathcal{IDK}$-$\mathcal{S}$ framework, as detailed below.

\subsection{Proposed Framework: $\mathcal{IDK}$-$\mathcal{S}$}
\label{sec:framework}

Given a data stream $\mathcal{S} = \{x_1,\dots, x_T, \dots\}\subset \mathbb{R}^d$, the proposed $\mathcal{IDK}$-$\mathcal{S}$ framework operates in two main phases: (1) an one-off initialization phase to initialize the IDK model in the first window of the data stream, and (2) a continuous incremental update phase that adapts the model to the evolving data stream while scoring every new instance.

\subsubsection{Initialization phase.} Let $\mathbf{X}_0 = \{x_1,\dots,x_{\omega}\}\subset\mathcal{S}$ denote the first data window in data stream $\mathcal{S}$, where $\omega$ is the fixed window size. 
$\mathcal{IDK}$-$\mathcal{S}$ first initializes the model on $\mathbf{X}_0$. As $\mathbf{X}_0$ can be treated as a given static data, this initialization phase is consistent with the IDK model~\cite{ting2021isolation}.

First, $\mathcal{IDK}$-$\mathcal{S}$ randomly selects $\psi$ ($2\leq\psi < \omega$) samples from $\mathbf{X}_0$ to get a subset $\mathcal{D}=\{s_1,\dots,s_\psi\}\subset\mathbf{X}_0$. For any sample $s\in\mathcal{D}$, its hypersphere partition $B_s$ centred at $s$ with radius $\tau_s$ is defined to be $B_s = \{x\in \mathbb{R}^d:\{\|x-s\|<\tau_s\}$,
where $\tau_s= \min_{y \in \mathcal{D}\setminus\{s\}} \|y- s\|$. Each hypersphere $B_s$ is a weak anomaly detector, defined by a mapping function $\phi_s: \mathbb{R}^d \to \{0, 1\}$ to output a score ($1$ for normal, $0$ for anomalous) for any instance $x\in \mathbb{R}^d$. As in IDK, $\phi_s(x)=1$ if and only if $x\in B_s$ and $x$ is closest to $s$ compared to other samples in $\mathcal{D}$, i.e., 
\begin{equation}
    \phi_s(x) = \mathds{1}(x\in B_s, \|x - s\| = \min_{y\in\mathcal{D}}\|x-y\|),
\label{eq:featuremap}
\end{equation}
where $\mathds{1}(\cdot)$ is the indicator function. 

Then repeat the above process $t$ times, $\mathcal{IDK}$-$\mathcal{S}$ gets a set of $t$ subsets $\{\mathcal{D}_i\}_{i=1}^{t}$ and a set of $\psi t$ detectors $\Phi_0 = \{\phi_{s_j}\}_{j=1}^{\psi t}$. For any $x\in\mathbb{R}^d$, its feature map is defined as 
\begin{equation}
    \Phi_0(x) = [\phi_{s_1}(x),\dots,\phi_{s_{\psi t}}(x)],
\label{eq:Phi}
\end{equation}
where $\Phi_0(x)\in\{0,1\}^{\psi \times t}$. 

The normal score, as computed in IDK, is defined as
\begin{equation}
    \text{score}(x) = \frac{1}{t} \langle \Phi_0(x), \widehat{\Phi}(\mathcal{P}_{\mathbf{X}_0}) \rangle,
\label{eq:score}
\end{equation}
where $\widehat{\Phi}(\mathcal{P}_{\mathbf{X}_0})=\frac{1}{\omega}\sum_{y\in\mathbf{X}_0}\Phi(y)$ is the feature mean map that is used to represent the overall data distribution of $\mathbf{X}_0$ in a Hilbert space. To avoid a costly summation over the entire window each time, $\mathcal{IDK}$-$\mathcal{S}$ maintains a running sum of all feature vectors $\sum_{y\in\mathbf{X}_0}\Phi(y)$, abbreviated as $\Sigma_{\Phi_{0}}$.

Let the model constructed by $\mathcal{IDK}$-$\mathcal{S}$ on $\mathbf{X}_0$ be denoted as $\Theta_0=\{\Phi_0,\Sigma_{\Phi_{0}}\}$.
For any data instance, its feature map is computed based on $\Phi_0$ using Eq.(\ref{eq:featuremap}) and Eq.(\ref{eq:Phi}). Its normal score is then computed through the dot product with $\Sigma_{\Phi_{0}}$ using Eq.(\ref{eq:score}). 
The core of $\mathcal{IDK}$-$\mathcal{S}$ is to continuously update $\Phi_0$ and $\Sigma_{\Phi_{0}}$ through a sample replacement strategy to cope with data stream.
It is important to note that, similar to other window-based streaming anomaly detectors, the effectiveness of the initial model $\Theta_0$ is contingent on the representativeness of the first data window $\mathbf{X}_0$~\cite{leveni2024online,Liu2025Streaming}. This ``cold start" consideration is a common aspect of online learning systems, where the initial state is learned from a limited subset of the entire data stream.

\subsubsection{Incremental Update Phase.}

The incremental update phase is triggered when the data window slides from $\mathbf{X}_i$ to $\mathbf{X}_{i+1}$, by discarding the obsolete data batch and incorporating the new data batch. To adapt to this change without the high cost of full retraining, $\mathcal{IDK}$-$\mathcal{S}$ performs a targeted, three-step update (see Figure~\ref{fig:idk_s} for an illustrative example).

\textit{1. Updating samples and partitions}:
The core of the update refreshes the model's hypersphere partitions. First, $\mathcal{IDK}$-$\mathcal{S}$ identifies the set of sample points that were generated from the now-obsolete batch. These samples and their corresponding hypersphere partitions are removed from $\Phi_{i}$. Second, an equal number of new sample points are randomly selected from the new data batch. New partitions are generated based on these new samples, resulting in an updated set of model partitions $\Phi_{i+1}$.

\textit{2. Updating the feature map}:
With the new set of partitions, the model updates the feature map $\Phi_{i+1}(x)$ for every instance $x \in \mathbf{X}_{i+1}$. This update is highly efficient. For instances in the new data batch, $\mathcal{IDK}$-$\mathcal{S}$ computes their full feature maps against all partitions. For instances that remain in the window, $\mathcal{IDK}$-$\mathcal{S}$ only recompute the map entries corresponding to the updated partitions.

\textit{3. Computing the mean map and scoring}:
Finally, $\mathcal{IDK}$-$\mathcal{S}$ updates the sum of feature vectors from $\Sigma_{\Phi_i}$ to $\Sigma_{\Phi_{i+1}}$. This is achieved by subtracting the feature vectors from obsolete instances, adding the feature vectors of new instances, and incorporating the feature vector changes for persistent instances. The updated model is therefore $\Theta_{i+1} = \{\Phi_{i+1}, \Sigma_{\Phi_{i+1}}\}$. For scoring the new instances, their normal scores are computed using the updated feature mean map, which is calculated on-the-fly as $\widehat{\Phi}(\mathcal{P}_{\mathbf{X}_{i+1}}) = \frac{1}{\omega}\Sigma_{\Phi_{i+1}}$.

By continually replacing the detectors associated with obsolete samples with new detectors generated from incoming data, $\mathcal{IDK}$-$\mathcal{S}$ effectively adapts to the current data distribution. The following proposition demonstrates that our proposed incremental strategy preserves a key statistical property of the retraining-based model.

\begin{prop}[Sampling Distribution Equivalence]
\label{prop:equivalence}
Given a data window $\mathbf{X}_i$ of size $\omega$ at any time step $i\in\mathbb{N}$. Let $\mathbf{S}_i \subset \mathbf{X}_i$ be the set of $\psi$ unique sample points used to generate the model partitions $\Phi_i$. The probability of obtaining a specific set $\mathbf{S}_i$ using the full retraining-based IDK is identical to the probability of obtaining it using $\mathcal{IDK}$-$\mathcal{S}$.
\begin{equation}
    P_{\text{IDK}}(\mathbf{S}_i \mid \mathbf{X}_i) = P_{\mathcal{IDK}\text{-}\mathcal{S}}(\mathbf{S}_i \mid \mathbf{X}_i) = \frac{1}{\binom{\omega}{\psi}},
\end{equation}
where $\binom{\omega}{\psi}$ is the binomial coefficient, representing the total number of ways to choose $\psi$ samples from the $\omega$ instances. 
\end{prop}

Proposition~\ref{prop:equivalence} demonstrates that at any time step, the probability distribution of the sample sets under both strategies is identical. This is because the incremental update mechanism in $\mathcal{IDK}$-$\mathcal{S}$ employs a random sampling process consistent with IDK, ensuring that the total probability of any final sample set appearing remains constant and equal to the uniform distribution probability of the retraining strategy.
This theoretical guarantee is critical, as it distinguishes $\mathcal{IDK}$-$\mathcal{S}$ from many heuristic incremental methods that might trade model fidelity for speed~\cite{leveni2024online,na2018dilof}, proving that our efficiency gains are achieved without compromising the statistical effectiveness of the model.
The proof of Proposition~\ref{prop:equivalence} is provided in Appendix.

\subsection{Complexity Analysis}

The computational complexity is a crucial consideration in the streaming context, where data streams must be processed at high speed under low memory constraints. To quantify the efficiency of our proposed $\mathcal{IDK}$-$\mathcal{S}$, we analyze its time and space complexities in comparison to the retraining-based IDK strategy. The complexities are summarized in Table~\ref{tab:complexity}, with parameters defined as: window size $w$, update step size $l$, number of partitions $t$, hyperspheres per partition $\psi$, and dimensionality of the data instance $d$.

\subsubsection{Time Complexity.}

The primary advantage of $\mathcal{IDK}$-$\mathcal{S}$ lies in its temporal efficiency. The time complexity of $\mathcal{IDK}$-$\mathcal{S}$ for an update step is $\mathcal{O}(l\psi t)$. This is derived from a detailed breakdown of the operations in its update phase.

First, compute the partitions corresponding to the updated samples. Let $m$ denote the number of sample replacements in one update step. On average, $m = \frac{l}{\omega} \psi t$ samples are replaced. Computing the radii for the corresponding partitions incurs a cost of $\mathcal{O}(m \cdot \psi)$, i.e., $\mathcal{O}(\frac{l}{\omega} \psi^2 t)$. Second, update the feature map. For instances that remain in the window, their feature maps are updated according to the new partitions. The complexity of this operation is $\mathcal{O}(m\cdot (\omega-l) )$, which simplifies to approximately $\mathcal{O}(l\psi t)$ (since $l \ll w$, the term $(\omega-l)/\omega$ is close to 1). For instances in the new data batch, computing their full feature maps against all partitions requires a time complexity of $\mathcal{O}(l\psi t)$. Third, score the new instance. Calculating the dot product of each new instance and the feature mean map requires time complexity $\mathcal{O}(l\psi t)$.

Therefore, the overall time complexity of $\mathcal{IDK}$-$\mathcal{S}$ is $\mathcal{O}(\frac{l}{\omega} \psi^2 t+l \psi t)$.
As the sampling size is smaller than the window size, i.e., $\psi < \omega$, $\mathcal{IDK}$-$\mathcal{S}$ has an overall time complexity of $\mathcal{O}(l\psi t)$, as the second term dominates the expression.
In contrast, the retraining-based IDK model needs to compute the feature maps of all $\omega$ instances via all $\psi t$ detectors, resulting in a time complexity of $\mathcal{O}(\omega \psi t)$.
Given that $l \ll w$, the incremental strategy of $\mathcal{IDK}$-$\mathcal{S}$ is more efficient and suitable for real-time applications.

\begin{table}[t]
\centering
\captionsetup{skip=5pt}

\renewcommand{\arraystretch}{1}
\begin{tabular}{@{}ccc@{}}
\toprule
 & Time & Space \\ \midrule
$\mathcal{IDK}$-$\mathcal{S}$ & $\mathcal{O}(l \psi t)$ & $\mathcal{O}(\omega d + \psi t d+  \omega \psi t)$ \\
Retraining-based IDK & $\mathcal{O}(\omega \psi t)$ & $\mathcal{O}(\omega d + \psi t d+  \omega \psi t)$ \\ \bottomrule
\end{tabular}
\caption{Complexity comparison of $\mathcal{IDK}$-$\mathcal{S}$ vs. IDK.}

\label{tab:complexity}
\end{table}

\subsubsection{Space Complexity.}
In terms of memory, both the incremental $\mathcal{IDK}$-$\mathcal{S}$ and the retraining-based IDK have the same space complexity of $\mathcal{O}(\omega d+ \psi t d+  \omega \psi t)$. 
This is because both of them need to store the data in the current window $\mathcal{O}(\omega d)$, all samples $\mathcal{O}(\psi t d)$, and the feature maps of data instances in current window $\mathcal{O}(\omega\psi t)$. Crucially, for $\mathcal{IDK}$-$\mathcal{S}$, this memory requirement is dependent on the window size $\omega$, not the total length of the data stream, making it viable for processing potentially infinite data streams.

\section{Experiments}

In this section, we compare $\mathcal{IDK}$-$\mathcal{S}$ with state-of-the-art methods on large-scale anomaly detection benchmarks, 
including both stationary and non-stationary distributions.

\subsection{Datasets} 

Following oIFOR~\cite{leveni2024online}, we use eleven large anomaly detection benchmarks, including the cardinality $n$ ranging from thousands to hundreds of thousands, dimensionality $d$ ranging from 3-48, and anomaly proportions ranging from 0.03\% to 32\%, to comprehensively test the performance of different methods on stationary data streams. Each stationary dataset is first randomly shuffled and then used for algorithm execution. 
We also use the \texttt{INSECTS} dataset~\cite{frittoli2023nonparametric,stucchi2023kernel}, which contains five real changes caused by temperature modifications that affect the insects’ flying behavior, 
to test the performance on non-stationary data streams~\cite{leveni2024online}. 
Moreover, we use the \texttt{TwoCluster} dataset shown in Figure~\ref{fig:f1} to test the performance on non-stationary data streams.
Different from the sudden concept drift of INSECTS, the concept drift of \texttt{TwoCluster} occurs in a more gradual form. The properties of datasets used in our experiments are outlined in Table~\ref{tab:dataset_properties}.

\begin{table}[t]
\centering
\captionsetup{skip=5pt}
\resizebox{0.9\columnwidth}{!}{%
\begin{tabular}{@{}clrrr@{}} %
\toprule
& Dataset & \multicolumn{1}{c}{$n$} & \multicolumn{1}{c}{$d$} & \% of anomalies \\
\midrule
\multirow{11}{*}{\rotatebox{90}{\textbf{Stationary}}} & \texttt{Donors} & 619,326 & 10 & 5.90 \\
& \texttt{Http} & 567,497 & 3 & 0.40 \\
& \texttt{ForestCover} & 286,048 & 10 & 0.90 \\
& \texttt{Fraud} & 284,807 & 29 & 0.17 \\
& \texttt{Mulcross} & 262,144 & 4 & 10.00 \\
& \texttt{Smtp} & 95,156 & 3 & 0.03 \\
& \texttt{Shuttle} & 49,097 & 9 & 7.00 \\
& \texttt{Mammography} & 11,183 & 6 & 2.00 \\
& \texttt{NYC\_taxi\_single} & 10,273 & 48 & 5.20 \\
& \texttt{Annthyroid} & 6,832 & 6 & 7.00 \\
& \texttt{Satellite} & 6,435 & 36 & 31.64 \\
\midrule
\multirow{2}{*}{\rotatebox{90}{\textbf{Non}}} & \texttt{INSECTS} & 212,514 & 33 & 5.50  \\
& \texttt{TwoCluster} & 1,000,000 & 2 & 5.00 \\
\bottomrule
\end{tabular}}
\caption{Datasets properties.}
\label{tab:dataset_properties}
\end{table}

\subsection{Competing Methods and Setups}

We compare the performance of $\mathcal{IDK}$-$\mathcal{S}$ against five baselines, which represent various research lines to extend foundational offline detectors to the streaming data paradigm. These competitors include two adaptations of the iForest detector, SiForest~\cite{Liu2025Streaming} and oIFOR~\cite{leveni2024online}; DILOF~\cite{na2018dilof}, which extends the density-based LOF detector; CPOD~\cite{tran2020real}, which extends the distance-based detector; and Memstream~\cite{bhatia2022memstream}, which combines a deep autoencoder with kNN for anomaly detection. It is worth noting that only Memstream runs on GPU in our experiment.

For all competing methods, we carefully tuned their parameters according to the suggestions in their papers. Following oIFOR~\cite{leveni2024online}, we set the window size $\omega=2048$ and the update step size $l=100$ for all methods.
Notably, for a fair comparison, the number of ensembles was set to the default $t=100$ for the ensemble methods, including $\mathcal{IDK}$-$\mathcal{S}$, SiForest and oIFOR. For $\mathcal{IDK}$-$\mathcal{S}$, the number of samples $\psi$ was searched in $[2^1,2^2,\dots,2^6]$. Each method was executed 20 times on all datasets. 
We adopted the widely used ROC AUC score and runtime (in seconds) to evaluate the effectiveness and efficiency of the methods~\cite{huang2005using}. All experiments are conducted on the same Linux CPU machine: AMD 128-core CPU with each core running at 2 GHz and 1T GB RAM. 

\subsection{Results}

\subsubsection{Detection Effectiveness.} In Table~\ref{tab:roc_auc_times}, we show the average AUC scores for each algorithm on all datasets. The results show that $\mathcal{IDK}$-$\mathcal{S}$ achieves the best AUC score in the most datasets, with the only exception of the \texttt{Shuttle} dataset. Interestingly, this result on \texttt{Shuttle} is consistent with the observation from existing research of the offline detectors, where the IDK (0.98) was also slightly outperformed by iForest (0.99) on this specific dataset, while IDK was superior on other datasets, including \texttt{Smtp} and \texttt{ForestCover}~\cite{ting2021isolation}. This observation is a reflection of our core argument that the performance of a streaming detector is fundamentally governed by the capabilities of its underlying offline base. The superiority of IDK is due to its use of a data-dependent kernel, which has been demonstrated to be more effective than the mechanisms in traditional detectors like iForest and LOF in most offline cases. This inherent advantage carries over directly to the streaming context. The proposed $\mathcal{IDK}$-$\mathcal{S}$ achieves the best mean AUC of 0.876, while its closest competitor, SiForest, obtains a mean of 0.853, showing a clear performance gap. 

\subsubsection{Adaptability to Concept Drift.} 
\begin{figure}[h]
    \centering
    \begin{subfigure}[b]{1\columnwidth}
        \includegraphics[width=0.96\columnwidth]{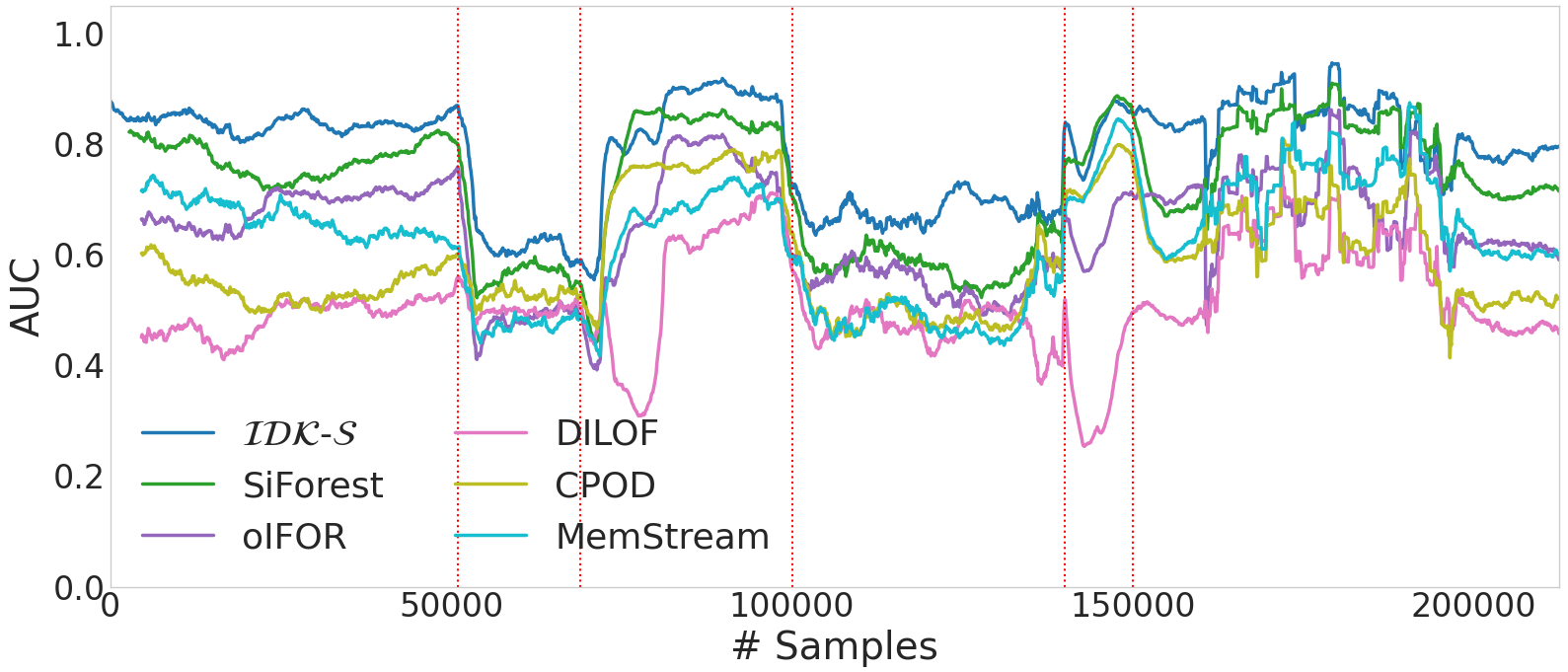}
        \caption{\texttt{INSECTS} dataset (abrupt change).}
        \label{fig:ins}
    \end{subfigure}
    \hfill 
    \begin{subfigure}[b]{1\columnwidth}
        \includegraphics[width=0.975\columnwidth]{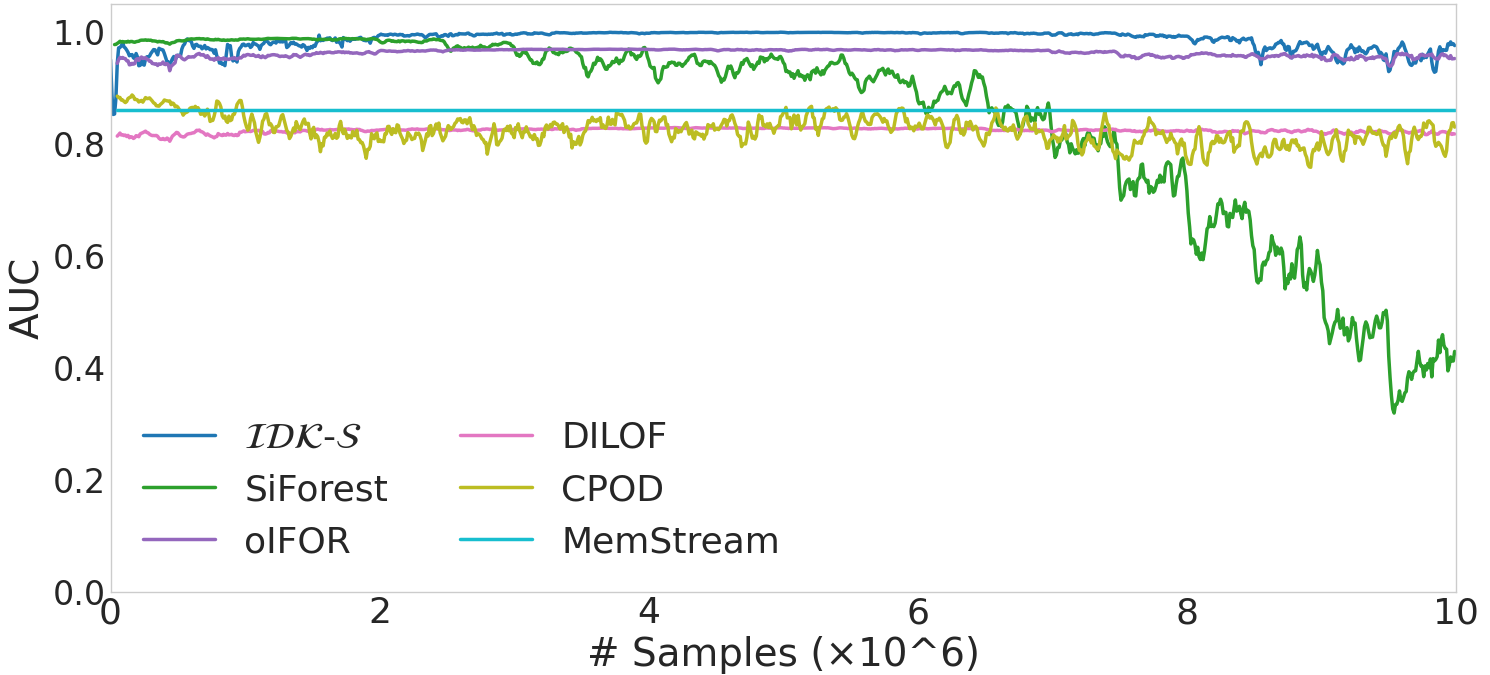}
        \caption{\texttt{TwoCluster} dataset (gradual change).}
        \label{fig:twoc}
    \end{subfigure}
    \caption{The performance of methods on non-stationary datasets where $\mathcal{P}_{N}$ and $\mathcal{P}_{A}$ change over time.}
    \label{fig:non_stationary}
\end{figure}

Figure~\ref{fig:non_stationary} investigates the adaptability of the methods on two non-stationary datasets. To assess instantaneous performance, we computed the AUC score at each time instant $T$ using normal scores within a sliding evaluation interval $\{x_{T-\omega},\dots,x_{T+\omega}\}$ of size $2\omega$, centred at $x_{T}$. 
The length of this interval was chosen to guarantee that the resulting curves exhibit a satisfactory degree of smoothness.
Overall, $\mathcal{IDK}$-$\mathcal{S}$ achieves the best performance compared to existing methods.
On the \texttt{INSECTS} dataset, which features abrupt concept drifts (Figure~\ref{fig:non_stationary}a), all methods show temporary but significant drops in AUC scores after each drift, underscoring a common challenge for current streaming detectors. In contrast, on the \texttt{TwoCluster} dataset with gradual drifts (Figure~\ref{fig:non_stationary}b), $\mathcal{IDK}$-$\mathcal{S}$ and most other methods maintain stable performance. The notable exception is SiForest, whose accuracy degrades significantly as the distribution evolves. In our experiments, we found that SiForest sometimes retains obsolete data for too long due to its reservoir sampling update mechanism, preventing the model from adapting to the current distribution over time. 

\subsubsection{Computational Efficiency.} 
Beyond superior detection performance, $\mathcal{IDK}$-$\mathcal{S}$ shows exceptional time efficiency, a critical requirement for real-world streaming applications. As detailed in Table~\ref{tab:roc_auc_times}, $\mathcal{IDK}$-$\mathcal{S}$ consistently records the lowest runtimes across all competing methods. This efficiency is also clearly evident even when compared to existing fastest baseline, oIFOR. On the six largest datasets, $\mathcal{IDK}$-$\mathcal{S}$ shows a speed-up of at least 5x. The runtime gap widens further on several of the smaller datasets (e.g., \texttt{Shuttle}, \texttt{NYC\_taxi\_single}, and \texttt{Satellite}), where the runtime of $\mathcal{IDK}$-$\mathcal{S}$ is reduced by a factor approaching or exceeding an order of magnitude.
This superior time efficiency of $\mathcal{IDK}$-$\mathcal{S}$ is due to three key factors. First, its base detector IDK has linear-time complexity, making it inherently faster than methods based on LOF and kNN which often have quadratic complexity~\cite{ting2021ikde}. 
Second, our incremental update strategy avoids costly full retraining, which decouples the update complexity from the overall window size and makes it dependent only on the much smaller update step size.
Third, compared to the iForest-based methods including SiForest and oIFOR with the same number of ensembles ($t$), $\mathcal{IDK}$-$\mathcal{S}$ achieves its optimal performance with a significantly smaller number of samples ($\psi$), which we discuss in detail in the following parameter analysis.

\begin{table*}[t]
\centering
\captionsetup{skip=5pt}
\resizebox{\textwidth}{!}{
\begin{tabular}{@{}c l cccccc cccccc@{}}
\toprule
& & \multicolumn{6}{c}{AUC ($\uparrow$)} & \multicolumn{6}{c}{TIME (s) ($\downarrow$)} \\
\cmidrule(lr){3-8} \cmidrule(lr){9-14}
& Dataset & $\mathcal{IDK}$-$\mathcal{S}$ & SiForest & oIFOR & DILOF & CPOD & Memstream & $\mathcal{IDK}$-$\mathcal{S}$ & SiForest & oIFOR & DILOF & CPOD & Memstream \\
\midrule
\multirow{11}{*}{\rotatebox{90}{\textbf{Stationary}}} & \texttt{Donors} & \textbf{0.802} & 0.780 & {0.796} & 0.673 & 0.603 & 0.761 & \textbf{57} & 694 & {301} & 1492 & 4227 & 604 \\
& \texttt{Http} & \textbf{0.999} & \textbf{0.999} & 0.996 & 0.732 & 0.698 & 0.998 & \textbf{29} & 426 & {183} & 633 & 1492 & 439 \\
& \texttt{ForestCover} & \textbf{0.926} & 0.907 & 0.891 & 0.703 & 0.729 & 0.885 & \textbf{19} & 315 & {115} & 578 & 1619 & 294 \\
& \texttt{Fraud} & \textbf{0.962} & 0.941 & 0.931 & 0.897 & 0.874 & 0.936 & \textbf{21} & 378 & {177} & 591 & 1584 & 345 \\
& \texttt{Mulcross} & \textbf{0.998} & 0.949 & {0.994} & 0.800 & 0.810 & 0.981 & \textbf{17} & 289 & {87} & 443 & 1136 & 251 \\
& \texttt{Smtp} & \textbf{0.911} & 0.894 & 0.863 & 0.854 & 0.851 & 0.904 & \textbf{6} & 203 & {31} & 196 & 559 & 179 \\
& \texttt{Shuttle} & 0.976 & \textbf{0.996} & 0.990 & 0.912 & 0.902 & 0.973 & \textbf{2} & 169 & {19} & 128 & 413 & 155 \\
& \texttt{Mammography} & \textbf{0.866} & 0.840 & 0.859 & 0.829 & 0.803 & 0.842 & \textbf{0.7} & 51 & {4} & 57 & 128 & 43 \\
& \texttt{NYC\_taxi\_single} & \textbf{0.732} & 0.719 & 0.565 & 0.507 & 0.491 & 0.702 & \textbf{1} & 102 & {11} & 98 & 266 & 79 \\
& \texttt{Annthyroid} & \textbf{0.752} & 0.713 & 0.678 & 0.619 & 0.608 & 0.709 & \textbf{0.7} & 27 & {3} & 39 & 85 & 3 \\
& \texttt{Satellite} & \textbf{0.726} & 0.699 & 0.667 & 0.477 & 0.636 & 0.678 & \textbf{0.5} & 42 & {5} & 36 & 24 & 4 \\
\midrule
\multirow{2}{*}{\rotatebox{90}{\textbf{Non}}} & \texttt{INSECTS} & \textbf{0.783} & 0.752 & 0.718 & 0.596 & 0.614 & 0.665 & \textbf{18} & 713 & 368 & 1447 & 4531 & 734 \\
& \texttt{TwoCluster} & \textbf{0.958} & 0.896 & 0.940 & 0.885 & 0.829 & 0.904 & \textbf{124} & 2194 & 1528 & 4419 & 9473 & 2366 \\
\midrule
\multicolumn{2}{c}{Mean} & \textbf{0.876} & 0.853 & 0.838 & 0.730 & 0.727 & 0.841 & \textbf{27} & 432 & 218 & 781 & 1916 & 423 \\
\multicolumn{2}{c}{Improvement (\%) / Speedup (\text{x})} & - & 2.2\% & 4.5\% & 20.0\% & 20.4\% & 4.2\% & - & 16\text{x} & 8\text{x} & 29\text{x} & 71\text{x} & 16\text{x} \\
\bottomrule
\end{tabular}%
}
\caption{The average AUC scores and runtimes. The best row-wise results are highlighted in bold.}
\label{tab:roc_auc_times}
\end{table*}

\begin{figure}[t]
    \centering
    \includegraphics[width=0.96\columnwidth]{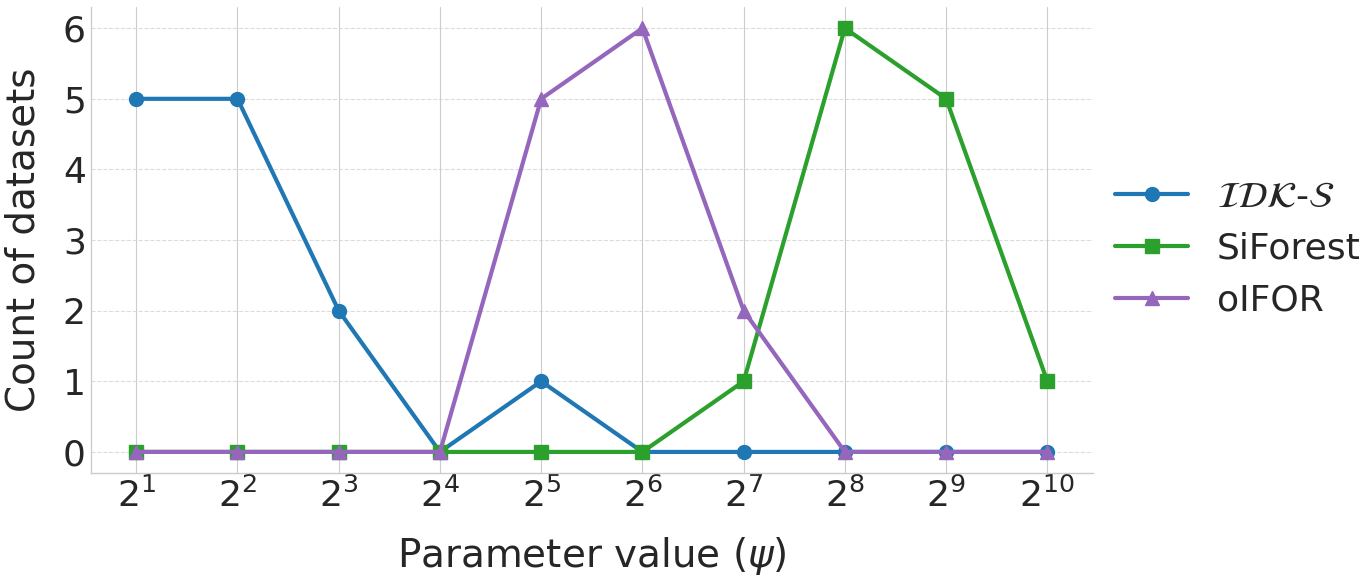} 
    \caption{Distribution of the optimal sample size $\psi$.}
    \label{fig:para}
\end{figure}

\begin{table}[t]
\centering

\resizebox{0.45\textwidth}{!}{%
\begin{tabular}{@{}llcc@{}}
\toprule
Type & Method & Average AUC & Average time (s) \\
\midrule
\multirow{2}{*}{Online} & $\mathcal{IDK}$-$\mathcal{S}$ & 0.877 & 14 \\
 & Retraining-based IDK & 0.879 & 352 \\
\midrule 
Offline & IDK & 0.898 & 8 \\
\bottomrule
\end{tabular}}
\label{tab:idk_variants_comparison}
\caption{Performance comparison of IDK variants.}
\end{table}

\subsection{Parameter Analysis}

To further understand the efficiency of $\mathcal{IDK}$-$\mathcal{S}$, we analyze its sensitivity to the number of samples, $\psi$, a key parameter that influences both performance and computational cost. We compare $\mathcal{IDK}$-$\mathcal{S}$ with the iForest-based ensemble methods, SiForest and oIFOR, which have analogous parameters (\textit{reservoir\_samples} and \textit{max\_leaf\_samples}, respectively). For comparison purposes, we set the number of ensemble estimators to $t=100$. Figure~\ref{fig:para} shows that the optimal sample size $\psi$ of $\mathcal{IDK}$-$\mathcal{S}$ is concentrated in the range of $[2,8]$, while that of oIFOR is $[32,128]$ and that of SiForest is $[256,512]$.
This finding highlights a fundamental advantage of our approach. The ability of $\mathcal{IDK}$-$\mathcal{S}$ to operate effectively with a much smaller $\psi$ is a key contributor to its superior time efficiency, as the complexity of both generating partitions and scoring instances is directly dependent on this parameter. It suggests that the data-dependent kernel of IDK is more sample-efficient in capturing the data distribution compared to the random space-partitioning mechanism of iForest.

\subsection{Comparison with IDK Variants}

We compare $\mathcal{IDK}$-$\mathcal{S}$ with its two variants. Table 4 summarizes their average performance on stationary datasets. The offline IDK, which processes each dataset as a single static batch, serves as a non-streaming performance reference. 
The results show that $\mathcal{IDK}$-$\mathcal{S}$ achieves results comparable to retraining-based IDK while being more than 20x faster, demonstrating the effectiveness of our approach. 

\section{Conclusion}

We introduce $\mathcal{IDK}$-$\mathcal{S}$, a novel streaming anomaly detector that successfully adapts the powerful Isolation Distributional Kernel (IDK) to address the limitations of existing methods which are based on weaker offline detectors. Our lightweight incremental update strategy is not an approximation; it maintains statistical equivalence to full retraining while significantly reducing the computational complexity of IDK. Extensive experiments demonstrated that $\mathcal{IDK}$-$\mathcal{S}$ achieves both state-of-the-art accuracy and a speed-up of up to an order of magnitude over existing methods. This dual advantage is driven by the superior efficiency of its underlying data-dependent kernel. As a fast and accurate approach, $\mathcal{IDK}$-$\mathcal{S}$ is well-suited for real-world anomaly detection in evolving high-volume data streams. A future work would explore the challenge of improving robustness to abrupt concept drifts, 
a common weakness of current methods.

\newpage
\section*{Acknowledgements}
 This work is supported by National Natural Science Foundation of China (Grant No. W2531050 and 92470116).

\bibstyle{aaai2026.bst}
\bibliography{aaai2026}

\newpage

\end{document}